\documentclass{IISERB}

\makenomenclature

\include{commands/commands}

\usepackage{etaremune}

\newcommand{\thesistitle}{Privacy-Preserving Machine Learning Models}
\newcommand{\studentname}{Aditya Mishra}
\newcommand{\studentrollno}{21013}
\newcommand{\advisorname}{Dr. Haroon Lone}
\newcommand{\subject}{Electrical Engineering and Computer Science}
\newcommand{\thesisdate}{April, 2025}

\begin{document}

\begin{titlepage}
\maketitle
\end{titlepage}

\pagebreak
\thispagestyle{firstpage}
\vspace*{1.5cm}
\begin{center}\Large{\textbf{CERTIFICATE}}\end{center}
\vspace{0.5cm}

This is to certify that {\bf \studentname}, BS (\subject), has worked on the project entitled {\bf `\thesistitle'} under my supervision and guidance. The content of this report is original and has not been submitted elsewhere for the award of any academic or professional degree.

\vspace{7em}

\textbf{\thesisdate \hfill \advisorname \\ IISER Bhopal} \null \hfill \textit{Assistant Professor, EECS Department} \\ \null \hfill \textit{IISER Bhopal}

\addcontentsline{toc}{chapter}{Certificate}
\newpage  
\addcontentsline{toc}{chapter}{Academic Integrity and Copyright Disclaimer}

\begin{center}
{\textbf{\Large{ACADEMIC INTEGRITY AND COPYRIGHT DISCLAIMER}}}
 \end{center}
I hereby declare that this thesis is my own work and, to the best of my knowledge, it contains no materials previously published or written by any other person, or substantial proportions of material which have been accepted for the award of any other degree or diploma at IISER Bhopal or any other educational institution, except where due acknowledgement is made in the thesis. \\ \\
I certify that all copyrighted material incorporated into this thesis is in compliance with the Indian Copyright (Amendment) Act, 2012 and that I have received written permission from the copyright owners for my use of their work, which is beyond the scope of the law. I agree to indemnify and save harmless IISER Bhopal from any and all claims that may be asserted or that may arise from any copyright violation.

\vspace{2cm}
\parbox{0.7\textwidth}{ 
	\flushleft{\thesisdate\\IISER Bhopal}
}
\hfill 
\parbox{0.3\textwidth}{ 
\mbox{\textbf{\studentname}}
}
\newpage
\addcontentsline{toc}{chapter}{Acknowledgement}

\begin{center}
{\textbf{\large{ACKNOWLEDGEMENT}}}
 \end{center}

I would like to express my sincere gratitude to Dr. Haroon Lone for his invaluable support throughout my research and for being my supervisor. His expertise and insights have been instrumental in my research. My heartfelt thanks go to my parents - Dr. Satanand Mishra and Mrs. Snehlata Mishra, my elder brother, Aayush, and my other family members who have always been there for me, providing me with their unwavering support and encouragement. Without their constant love and support, I could not have accomplished this. I would like to thank Riya, Baij, Vishal, residents of fifth-floor of Hostel-7, and
all my friends for their unwavering support throughout my journey at IISER Bhopal. Thank you all for your contributions to my academic journey and the completion of my thesis.

\vspace{2cm}

\hfill \textbf{\studentname}

\newpage 
\addcontentsline{toc}{chapter}{Abstract}

\begin{center}\Large{\textbf{ABSTRACT}}\end{center}

In academic settings, the demanding environment often forces students to prioritize academic performance over their physical well-being. Moreover, privacy concerns and the inherent risk of data breaches hinder the deployment of traditional machine learning techniques for addressing these health challenges. In this study, we introduce \textbf{RiM: Record, Improve, and Maintain}, a mobile application which incorporates a novel personalized machine learning framework that leverages federated learning to enhance students’ physical well-being by analyzing their lifestyle habits.
\newline
\par

Our approach involves pre-training a multilayer perceptron (MLP) model on a large-scale simulated dataset to generate personalized recommendations. Subsequently, we employ federated learning to fine-tune the model using data from IISER Bhopal students, thereby ensuring its applicability in real-world scenarios. The federated learning approach guarantees differential privacy by exclusively sharing model weights rather than raw data. Experimental results show that the FedAvg–based RiM model achieves an average accuracy of 60.71\% and a mean absolute error of 0.91—outperforming the FedPer variant (average accuracy 46.34\%, MAE 1.19)—thereby demonstrating its efficacy in predicting lifestyle deficits under privacy‐preserving constraints.

\let\cleardoublepage\clearpage
\newpage
\addcontentsline{toc}{chapter}{Table of Contents}
\tableofcontents
\newpage


\addcontentsline{toc}{chapter}{List of Symbols and Abbreviations}%


\nomenclature{$\tau$}{Step Threshold}

\nomenclature{$\delta$}{Stride Length}

\nomenclature{RiM}{Record, Improve and Maintain}

\nomenclature{ML}{Machine Learning}

\nomenclature{MLP}{Multilayer Perceptron}

\nomenclature{MTL}{Multi-task Learning}

\nomenclature{FL}{Federated Learning}

\nomenclature{MAE}{Mean Absolute Error}

\nomenclature{FedAvg}{Federated Averaging}

\nomenclature{FedPer}{Federated Personalization}

\nomenclature{FedPer}{Federated Personalization}

\nomenclature{$w$}{Weights}

\nomenclature{$\tau_{j}$}{Deficit Threshold}

\nomenclature{$R$}{Composite Risk Score}

\nomenclature{$d$}{Deficit}

\nomenclature{$\theta$}{Risk Score Threshold}

\printnomenclature


\let\cleardoublepage\clearpage

\addcontentsline{toc}{chapter}{List of Figures}
\listoffigures
\let\cleardoublepage\clearpage
\addcontentsline{toc}{chapter}{List of Tables}
\listoftables

\pagestyle{fancy}
\clearpage
\pagenumbering{arabic}

\newpage


\chapter{Introduction}

\section{Background and Motivation}

The intense competitive environment in academic settings has driven students to prioritize their studies and academic goals, often at the expense of their physical well-being. The rise of technological distractions, such as video games and binge-watching, has further exacerbated this issue. Furthermore, privacy concerns and the risk of data breaches in traditional machine learning approaches have limited the potential of AI to address these challenges for students.
\newline \par

To address these issues, we present RiM: Record, Improve, and Maintain, a personalized federated learning-based model designed to enhance students' physical well-being by analyzing their lifestyle habits. By leveraging federated learning, RiM ensures differential privacy, guaranteeing that we have no access to the data collected from students, thus addressing both privacy and wellness concerns effectively. 

\section{Literature Review}

Recent meta‑analyses and randomized trials underscore the efficacy of digital interventions in promoting physical well-being. \cite{bi2024effectiveness} conducted a meta‑analysis of 15 studies targeting university students and found that personalized SMS reminders and smartphone applications led to significant increase in daily step counts, though effects on moderate‑to‑vigorous physical activity and sedentary behavior varied across studies. \cite{figueroa2024ratings} underscore the significance of mobile health applications in enhancing students' motivation, effectively promoting physical activity levels.  \cite{tong2022use} explored that mobile apps and fitness trackers are positively associated with increased physical activity during the pandemic. It highlighting the pivotal role digital tools play in supporting health-related behaviors. Furthermore, studies \cite{kim2023effectiveness, firth2024using} demonstrate the effectiveness of mobile phone–based physical activity programs in reducing symptoms of depression and perceived stress, as well as promoting healthier lifestyle choices among young individuals experiencing mental health conditions.
\newline \par

\textbf{Multi‑task learning} (MTL) is a paradigm that trains a single model to perform multiple related tasks by sharing representations, improving generalization especially when per‑task data are limited. The work by \cite{zhang2021survey} depicts that MTL mitigates overfitting and leverages inter‑task correlations. \cite{chih2024multitask} developed an MTL‑enhanced convolutional‑RNN architecture that jointly predicts sleep stages and heart‑rate variability from ECG and PPG data. Their model achieved the accuracy similar to single‑task baselines while using 75\% less input data and 7.5 times fewer parameters. In the clinical domain, \cite{ali2022multitask} introduced a multimodal LSTM‑based MTL model to predict both hospital length‑of‑stay (regression) and 30‑day readmission (classification) using wrist‑worn sensor data. Their joint model significantly outperformed separate single‑task models on both objectives. Together, these works illustrate MTL’s ability to handle concurrent goals under constrained data and compute budgets—an approach we adopt by fine‑tuning an MLP to jointly predict sleep and distance deficits from shared lifestyle features. 
\newline \par

The \textbf{federated learning} (FL) paradigm enables on‑device training by exchanging only model updates, thereby keeping each user’s data on their own device. \cite{mcmahan2017communication} formalized this approach with the FedAvg algorithm, showing that iterative averaging of locally‑computed updates can train deep networks across non‑IID mobile data while reducing communication rounds by 10-100 times while preserving data locality. Building formal privacy guarantees into FL, \cite{geyer2017differentially} proposed DP‑FedAvg. It applies per‑client gradient clipping and Gaussian noise to each update, achieving user‑level ($\epsilon,\delta$)‑differential privacy with only minor accuracy degradation given a sufficiently large cohort. Furthermore, \cite{zheng2021federated} developed a federated f‑differential privacy framework that leverages Gaussian differential privacy and tight composition analyses to offer both record‑level and group‑level privacy guarantees under federated settings. This further tightens the privacy‑utility trade‑off. To support research and real‑world deployment at scale, \cite{beutel2020flower} released Flower. It is an open-source platform that provides high‑level building blocks to prototype FL workflows quickly and works across many different devices and environments, from mobile phones to edge servers. Also, it scales seamlessly from single‑node simulations to millions of clients on real devices.
\newline \par

\section{Introduction to RiM}

The decentralized approach of training models by distributing the data on different clients and learning a shared model by aggregating locally computed weights is known as \textbf{federated learning}. This concept was first introduced by \cite{DBLP:journals/corr/McMahanMRA16}. It preserves the sensitive user information by sharing only the weights of the model and not the raw data, thus ensuring differential privacy.
\newline \par

The developed mobile application is designed to capture the user's daily step counts, distance traveled, sleep hours, and meal information. The pre-trained MLP model is then fine-tuned on the user's data and the updates are shared to the \textbf{Flower framework}\footnote{{\hypersetup{urlcolor=blue}\url{https://flower.ai}}}. Flower is an open-source framework designed to simplify and streamline federated learning tasks on a cluster of machines. This Python-based framework offers a user-friendly solution for training a wide range of models, including deep neural networks.  The shared weights are combined via \textbf{Federated Averaging} \cite{mcmahan2017communication} and \textbf{Federated Personalization} \cite{arivazhagan2019federated} to update the global model. Subsequently, the updated weights are sent to each client device. 
\newline \par

After fine-tuning, the MLP model predicts sleep deficit and distance deficit from the ideal range by using a variety of features, such as: user's height, weight, age, gender, sleep hours, distance traveled, and meal information. A combination of rule-based and ML-based approaches enables the model to learn the inter-parameter relations and provides the user with how changes in one aspect of the lifestyle can affect the other parameters of physical well-being. Furthermore, priority-based mechanism is incorporated in order to provide only relevant and most important recommendation.
\newline \par

We describe our data pipeline in Chapter 2: first the generation of a large‐scale synthetic dataset for pre‑training, then the fine‑tuning dataset collected via the RiM Android app and its processing. In Chapter 3, we present the RiM mobile application, detail the MLP architecture and rule‑based recommendation system, explain our use of FedAvg and FedPer for privacy‑preserving fine‑tuning, outline the evaluation metrics and key implementation settings. Chapter 4 reports our experimental findings—comparing accuracy and MAE for FedAvg versus FedPer, analyzing class‑imbalance and personalization effects, and illustrating trade‑offs. It is followed by an in‑depth discussion of the results. Finally, Chapter 5 summarizes our contributions, acknowledges the Android 13–only limitation, and proposes future work on broadening version support, on‑device model integration, and advanced bi‑level federated algorithms like Ditto.

\let\cleardoublepage\clearpage


\chapter{Data}

\section{Pre-training on Simulated Data}

The lack of availability of a real-world data set with the set of features taken into consideration in this study makes us use simulated data set for pre-training the MLP model. The data set is generated for each feature separately, as each feature can be approximated to follow different distribution in the real-world scenario. 
\newline \par

Creating a realistic synthetic dataset for the study requires assigning appropriate statistical distributions to each feature based on empirical data. Table \ref{table:sim_data} illustrates the distribution each feature follow.

{\footnotesize
\begin{table}[h!]
    \centering
    \begin{tabular}{p{5cm}cc}
        \toprule
        \textbf{Features} & \textbf{Distribution} & \textbf{References}  \\
        \midrule
        Step Count & Negative Binomial Distribution & \cite{tudor2004many}\\
        Distance Traveled & Log-Normal Distribution & \cite{brockmann2006scaling}\\
        Sleep Hours & Normal Distribution & \cite{lauderdale2008self} \\
        Meal Consumption & Bernoulli Distribution & \cite{DWYER2001798}\\
        Height & Normal Distribution & \cite{silventoinen2003determinants}\\
        Weight & Log-Normal Distribution & \cite{flegal2012prevalence}\\
        Age & Truncated Normal (Empirical Distribution) & \cite{UN2019}\\
        Gender & Bernoulli Distribution & \cite{WHO2018}\\
        \bottomrule
    \end{tabular}
    \caption{Probabilistic distribution followed by each feature in the simulated data.}
    \label{table:sim_data}
\end{table}
}

\begin{figure}
    \centering
    \includegraphics[width=0.7\textwidth]{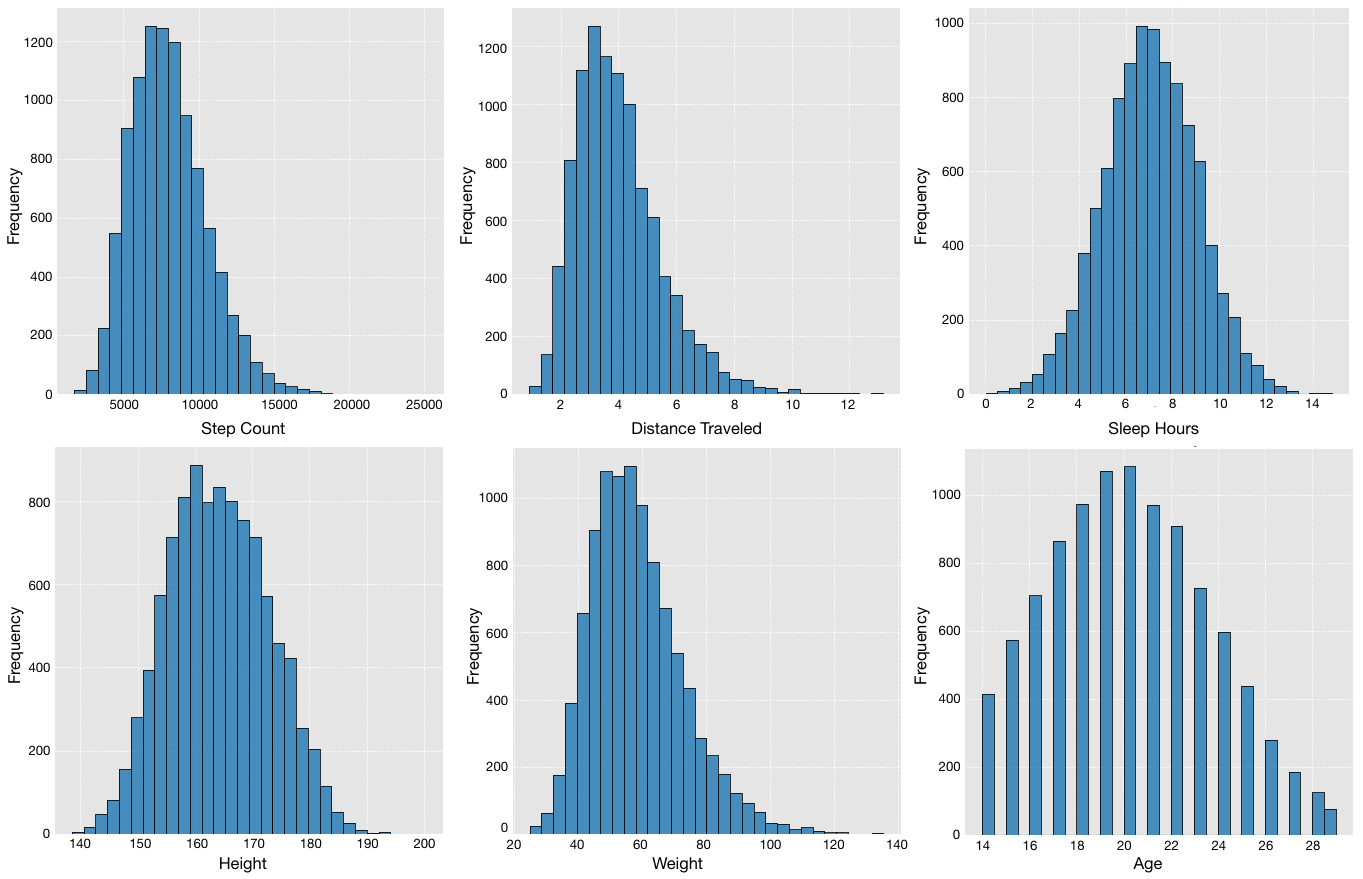}
    \caption{The figure illustrates the distributions of simulated features used for pre-training the MLP model. Simulated step counts follow a negative binomial distribution, while distance traveled and weight are drawn from log-normal distributions. In addition, meal consumption and gender are generated using Bernoulli distributions, and age is modeled based on an empirical distribution. The figure also illustrates the normal distributions for sleep hours and height.}
    \label{fig:sim1}
\end{figure}

\begin{figure}
    \centering
    \includegraphics[width=0.7\textwidth]{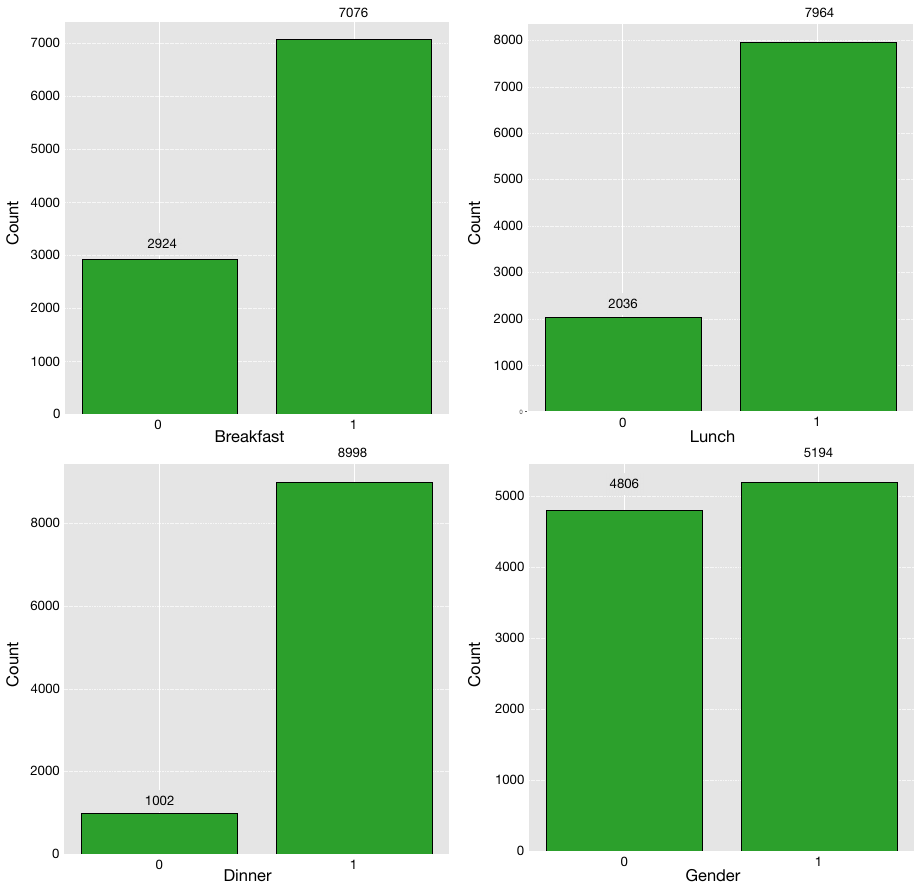}
    \caption{The figure presents the distribution of meal consumption—breakfast, lunch, and dinner—where a value of 1 indicates the meal was taken and 0 indicates it was skipped. It also shows the gender distribution within the simulated dataset. All variables follow a Bernoulli distribution.}
    \label{fig:sim2}
\end{figure}

\section{Fine-tuning on Real-world Data}

Through the developed mobile application, we collect the user's physical activity and demographic data locally. The model is fine-tuned on this data to adapt better to the real-world aspect. Table \ref{table:real_data} presents an example of the collected data. We collected user data over a 15-day period. An MLP model was fine-tuned using the data from the first 8 days and subsequently used to generate recommendations for the following 7 days. FedPer algorithm is used for fine-tuning the MLP model.
\newline \par

\begin{table}[h!]
\centering
\caption{The table depicts the collected data for a period of 1 week of a female user. \textsuperscript{\textdagger} represent that a feature is binary. }
\label{table:real_data}
\setlength{\tabcolsep}{5pt}
\resizebox{\textwidth}{!}{
\begin{tabular}{ccccccccccc} 
\toprule
\textbf{Date} & \textbf{Steps (\#)} & \textbf{Distance (km)} & \textbf{Sleep (hrs)} & \textbf{Breakfast\textsuperscript{\textdagger}} & \textbf{Lunch\textsuperscript{\textdagger}} & \textbf{Dinner\textsuperscript{\textdagger}} & \textbf{Age (yrs)} & \textbf{Height (cm)} & \textbf{Weight (kg)} & \textbf{Gender\textsuperscript{\textdagger}}\\
\midrule
2025-04-18 & 9657 & 4.83 & 8.04 & 0 & 1 & 1 & 22 & 165 & 59 & 0 \\
2025-04-19 & 8234 & 4.11 & 8.21 & 1 & 1 & 1 & 22 & 165 & 59 & 0 \\
2025-04-20 & 7698 & 3.84 & 7.83 & 0 & 0 & 1 & 22 & 165 & 59 & 0 \\
2025-04-21 & 11345 & 5.67 & 7.94 & 1 & 1 & 1 & 22 & 165 & 59 & 0 \\
2025-04-22 & 12876 & 6.43 & 6.54 & 1 & 1 & 1 & 22 & 165 & 59 & 0 \\
2025-04-23 & 6456 & 3.22 & 7.37 & 1 & 0 & 1 & 22 & 165 & 59 & 0 \\
2025-04-24 & 9825 & 4.91 & 6.18 & 1 & 1 & 0 & 22 & 165 & 59 & 0 \\
\bottomrule
\end{tabular}
}
\end{table}

Eighteen volunteers (14 male and 4 female) initially enrolled in the study and downloaded the mobile application. However, attrition and data-quality issues reduced the evaluable data to ten participants. The primary factors contributing to this reduction in evaluable datasets are as follows:

\begin{itemize}
    \item Volunteer withdrawals (n=4) occurred because of scheduling conflicts and competing personal or professional commitments.

    \item Technical failures (n=2) malfunctions of participants’ accelerometer sensors produced incomplete and corrupted datasets that failed to meet our predefined quality criteria.

    \item Insufficient engagement (n=2) was observed in participants who failed to meet the minimum interaction threshold (entering meal information), rendering their data unusable for reliable analysis.
\end{itemize}

Consequently, only the ten remaining datasets—each meeting compliance, completeness, and quality criteria—were included in the final analysis.
\newline \par

Since step count and distance exhibit a near‐perfect linear relationship (Pearson’s $r \approx 0.98$) \cite{tudor2004many}, we retain only the distance feature to mitigate multicollinearity and reduce dimensionality. Rather than passing height and weight separately, we compute the body mass index (BMI) to encapsulate overall physical fitness in a single metric—a practice shown to enhance physiological‐model performance in large‐scale cohort studies \cite{who_obesity_2000}. Breakfast consumption has been identified as the principal daily meal influencing metabolic health and cognitive performance, and thus is modeled as a distinct binary feature \cite{rampersaud2005breakfast}. To capture overall dietary patterns without inflating model complexity, we aggregate lunch and dinner into a single “meal” feature representing the total number of meals per day, consistent with approaches in nutritional epidemiology \cite{kant_graubard_2010}.
\newline \par

\let\cleardoublepage\clearpage


\chapter{Method}

\section{RiM: Mobile Application}

The onset of smartphones has revolutionized how we monitor our lifestyles and collect data effortlessly. In today’s technology-driven world, our devices remain with us at all times, providing continuous insights into our daily habits. Thus, we develop an android application that tracks and stores users’ lifestyle data locally, including physical activity, sleep duration, and dietary information. This real-world data plays a crucial role in fine-tuning our model to adapt to everyday scenarios.
\newline \par

We use \textbf{React Native}\footnote{{\hypersetup{urlcolor=blue}\url{https://reactnative.dev}}} to develop the application. It is an open-source UI software framework developed by Meta for creating mobile applications. The Android application exclusively utilizes accelerometer data to monitor and store parameters, ensuring computational efficiency and reduced battery consumption. Upon launch, the application requests the user's demographic details, such as height, weight, gender, and age. These details further help the MLP model tailor personalized recommendations. 
\newline \par

We subscribe to the phone's accelerometer for calculating the user's daily \textbf{step count} and \textbf{distance traveled}. Algorithm \ref{algo:step_counter} demonstrates how the step count is calculated. The algorithm continuously monitors accelerometer readings by calculating the magnitude of the acceleration vector from its \(x\), \(y\), and \(z\) components. It compares the change in magnitude from the previous reading against a predefined threshold ($\tau$). When this difference exceeds the threshold, a step is recorded. To prevent noisy data and  sensor fluctuations from triggering false step counts, a \textbf{debounce mechanism} is implemented. This mechanism ensures that a new step is only recorded if at least 300 milliseconds have passed since the last detected step. By imposing this time delay, the algorithm filters out rapid, minor changes that are unlikely to be actual steps, thereby enhancing the accuracy and reliability of the step detection process.
\newline \par

\begin{algorithm}[htbp]
\caption{Step Counter with Debounce}\label{algo:step_counter}
\begin{algorithmic}[1]
\State $lastMagnitude \gets 0$
\State $lastStepTime \gets 0$ \Comment{Timestamp of the last detected step in milliseconds}
\For{each accelerometer reading $(x, y, z)$}
    \State $magnitude \gets \sqrt{x^2 + y^2 + z^2}$
    \State $currentTime \gets$ current timestamp in milliseconds
    \If{$\lvert magnitude - lastMagnitude \rvert > \tau$ \textbf{and} $(currentTime - lastStepTime) > 300$}
        \State $lastStepTime \gets currentTime$
        \State $stepCount \gets stepCount + 1$
        \State $distance \gets distance + \delta$
        \State Update the last activity time with the current time
    \EndIf
    \State $lastMagnitude \gets magnitude$
\EndFor
\end{algorithmic}
\end{algorithm}

We record the user's daily \textbf{sleep hours} by monitoring the inactivity in the night. As the algorithm \ref{algo:sleep_tracking} illustrates, the sleep tracking algorithm adjusts sleep hours by combining two mechanisms. First, within the designated sleep window (10 PM to 10 AM), it calculates the time elapsed since the last recorded activity. If the inactivity period exceeds 2 hours, the user is marked as sleeping and sleep hours are incrementally increased at a rate of one minute. Second, if the user is flagged as sleeping and is then detected moving (between midnight and 10 AM), the algorithm compensates for lost sleep by adding 2 hours to the sleep total and resets the sleeping state. This dual approach ensures that short interruptions in activity do not skew sleep measurements, while still accurately capturing extended periods of inactivity as sleep.

\begin{algorithm}[htbp]
\caption{Sleep Hours Tracking}\label{algo:sleep_tracking}
\begin{algorithmic}[1]
    \State $now \gets$ current time
    \If{$now.\text{hour} \geq 22$ \textbf{or} $now.\text{hour} < 10$} \Comment{Check for sleep window}
        \State $timeSinceLastActivity \gets \dfrac{now - lastActivityTime}{1000 \times 60}$ \Comment{Convert to minutes}
        \If{$timeSinceLastActivity > 120$} \Comment{Inactivity for more than 2 hours}
            \If{$\lnot isSleeping$}
                \State $isSleeping \gets \text{true}$
            \EndIf
            \State $sleepHours \gets sleepHours + \dfrac{1}{60}$ \Comment{Increment sleep hours by 1 minute}
        \Else
            \State $currentTime \gets$ current time
            \If{$isSleeping$ \textbf{and} $0 \leq currentTime.\text{hour} < 10$}
                \State $sleepHours \gets sleepHours + 2$ \Comment{Add 2 hours to compensate lost sleep}
                \State $isSleeping \gets \text{false}$
            \EndIf
        \EndIf
    \EndIf
\end{algorithmic}
\end{algorithm}

To record the user's \textbf{meal information}, we create a drop-down input box with yes/no options that asks whether the user has taken their meal. To reduce user burden, a notification is scheduled at 21:45 hours to remind the user to enter their meal information once a day. This simple interface ensures that meal data is consistently captured without causing fatigue, while all other parameters are automatically tracked in the background, requiring no user input. For more details on the development of the Android application please visit GitHub\footnote{{\hypersetup{urlcolor=blue}\url{https://github.com/adityamishraaaa/RiM.git}}}.

\section{Machine Learning Frameworks}

\begin{figure}[htbp]
    \centering
    \includegraphics[width=0.9\textwidth]{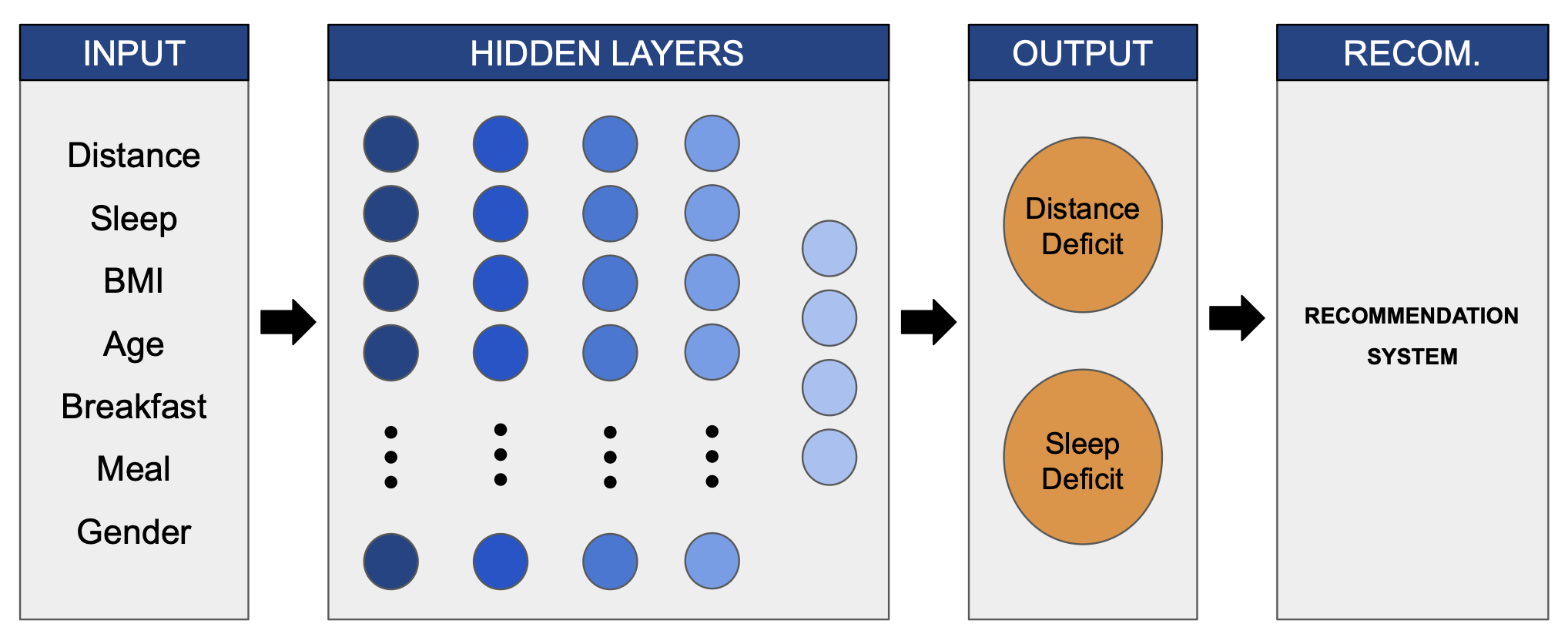}
    \caption{Architecture of the proposed MLP model, which consists of five hidden layers and outputs sleep and distance deficits that feed into the recommendation system to generate personalized recommendations.}
    \label{fig:mlp_model}
\end{figure}

The choice of a multi-layer perceptron (MLP) model is motivated by its ability to be fine-tuned on a relatively small dataset (7 days in our case), its low computational overhead, and ease of integration with Android applications. The proposed MLP architecture consists of five hidden layers. The first hidden layer contains 32 neurons and the number of neurons is halved in each subsequent layer, resulting in 4 neurons in the final hidden layer. Figure \ref{fig:mlp_model} illustrates the architecture of the proposed method.

We employ a two‐stage approach that (i) predicts lifestyle deficits via a pretrained MLP regression model and (ii) applies rule‐based severity and interaction checks to compute per‐parameter risk scores. Only the highest‐risk and most relevant recommendations are presented to the user.
\newline \par

\noindent Given a user feature vector
\[
\mathbf{x} = [\mathrm{distance},\,\mathrm{sleep},\,\mathrm{bmi},\,\mathrm{age},\,\mathrm{breakfast},\,\mathrm{meal},\,\mathrm{gender}],
\]
we first standardize:
\[
\tilde{\mathbf{x}} = \mathrm{StandardScaler}(\mathbf{x}).
\]
The MLP model \(f\) then predicts deficits
\[
\mathbf{d} = f(\tilde{\mathbf{x}})
= \bigl[d_{\text{sleep}},\,d_{\text{distance}}\bigr],
\]
where each deficit \(d_j\) is defined as
\[
d_j =
\begin{cases}
\;\; (\text{ideal}_{j,\min} - x_j), & x_j < \text{ideal}_{j,\min},\\[6pt]
-(x_j - \text{ideal}_{j,\max}), & x_j > \text{ideal}_{j,\max},\\[6pt]
\;\; 0, & \text{otherwise}.
\end{cases}
\]

\noindent For each primary parameter \(j\in\{\text{sleep},\,\text{distance}\}\), we only generate a recommendation if
\[
\lvert d_j\rvert > \tau_j,
\]
where \(\tau_j\) is a deficit threshold.  In that case, we assign
\begin{equation}\label{eq:param_risk}
r_j = w_j\,\lvert d_j\rvert,
\end{equation}
with weights \(w_{\text{sleep}}\) and \(w_{\text{distance}}\).  A corresponding message \(m_j\) (e.g.\ “Try to get more sleep”) is paired with each \(r_j\).
\newline \par


We also include rule‐based risks for meal skipping, abnormal BMI and inter-parameter interactions. Each rule contributes an additional risk score by the same formula \eqref{eq:param_risk}, with its own weight. Finally, we compute the composite risk score summing over all parameters and interactions:
\begin{equation}\label{eq:composite_risk}
R = \sum_{j} r_j,
\end{equation}
If \(R > R_{\text{high}}\), we prepend a high priority tag. To ensure conciseness, we filter out any \(r_j < \theta\). Sort the remaining risk–message pairs \(\{(r_j,m_j)\}\) in descending order of \(r_j\) and finally select the top \(N\) messages. If no \(r_j\) survives filtering, a generic \textbf{“Your lifestyle parameters are close to ideal. Keep it up!”} message is shown instead.
\newline \par

\begin{figure}[t]
    \centering
    \includegraphics[width=\textwidth]{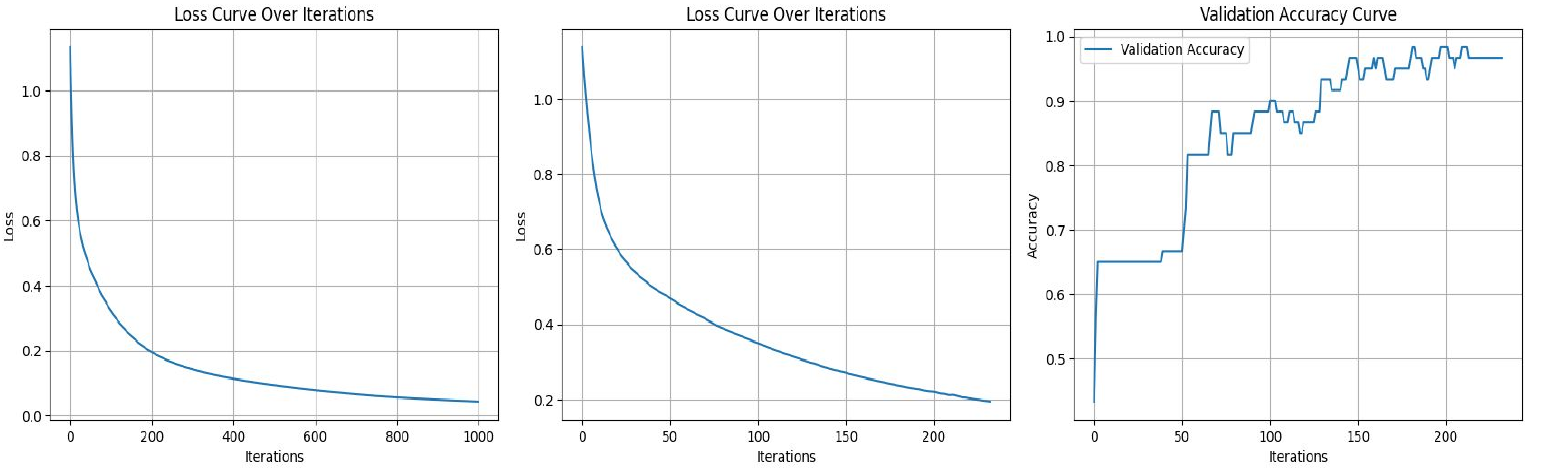}
    \caption{Training loss and validation accuracy curves plotted across pre‑training iterations of the MLP model.}
    \label{fig:loss}
\end{figure}

This design guarantees that only parameters with meaningful model‐predicted deficits or significant rule violations yield recommendations, that compound risks are captured via interaction rules, and that the user sees only the most pressing, relevant advice. The MLP model is pre-trained on simulated data using separate training and validation sets, along with an early stopping mechanism to prevent overfitting. Figure \ref{fig:loss} depicts the training curves. Pre-training allows the model to learn general patterns and representations. Furthermore, it helps to mitigate the limitations posed by the scarcity of real-world data.

\section{Federated Learning}

We use the Flower framework to implement federated learning. A central server is set up to load the pre-trained model and distribute its weights to individual clients, where each client represents a single user. Each client then fine-tunes the pre-trained model using its own real-world data. The data remains on the client side, ensuring that users' data stays isolated and does not interfere with other clients.
\newline \par

\begin{figure}[h!]
    \centering
    \includegraphics[width=0.9\textwidth]{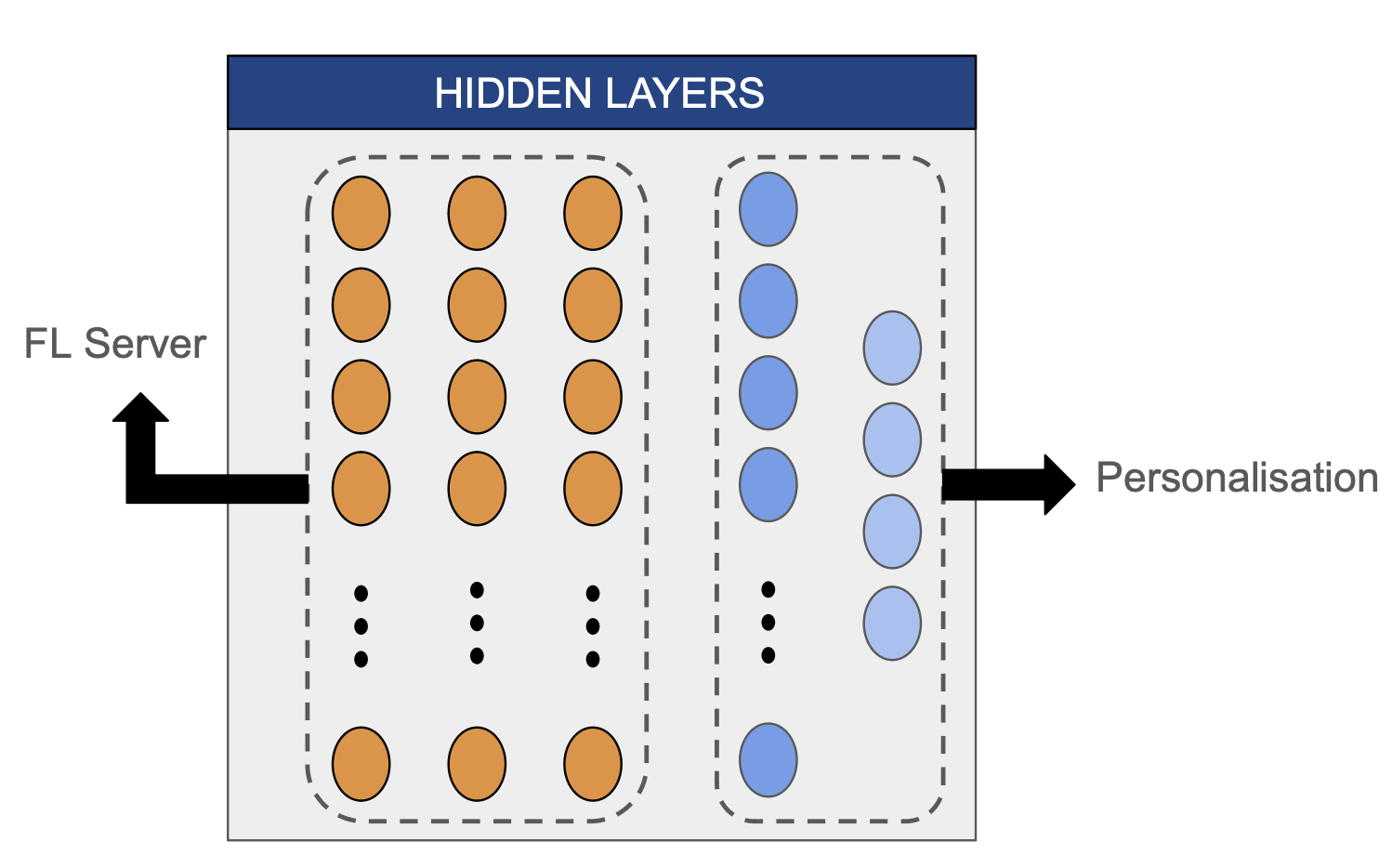}
    \caption{The figure illustrates the FedPer weight‐update process: orange neurons denote the shared root layers that are sent to the server averaged via FedAvg, while blue neurons denote the personal head layers that remain on the client device to preserve personalization.}
    \label{fig:fl_weight}
\end{figure}

We employ both FedAvg and FedPer algorithms to fine-tune the model. The FedAvg algorithm builds a global model by having each client train a local copy on its own data, then sending only the updated weights (not the data) back to a central server. The server aggregates these local updates by computing a weighted average to form a new global model. This global model is then redistributed to clients for the next round, repeating until convergence. FedPer splits the MLP model into a \textbf{shared root }(the first three layers) and \textbf{personal head} (the remaining  two layers), so that clients collaboratively train only the root while keeping their heads local. In each round, every client fine‑tunes the full model on its own data, then sends only the updated root weights to the server; the server averages these root updates (as in FedAvg) and broadcasts the new shared root back. This way, clients benefit from global feature learning but retain personalized output layers that capture local data heterogeneity. The FedPer fine-tuning process is illustrated by the Figure \ref{fig:fl_weight}. The orange neurons denote the shared root layers, while the blue neurons denote the personalized head layers.




\section{Evaluation Metrics}
We evaluated the performance of the models with two evaluation metrics – accuracy and mean absolute error (MAE).
\newline \par

\noindent{\textbf{Accuracy:}} The accuracy metric here measures the fraction of samples for which the predicted and true deficits share the same sign (positive, zero, or negative). In other words, it evaluates how often our model correctly predicts whether each deficit is above, equal to, or below zero.

\noindent Mathematically, if we have $n$ samples, true targets ${y}$ and predictions $\hat{y}$, then:

\begin{equation}\label{eqn:acc}
\mathrm{Accuracy}
= \frac{1}{n} \sum_{i=1}^{n}
\mathbf{1}\bigl(\mathrm{sign}(y_i) = \mathrm{sign}(\hat{y}_i)\bigr)
\end{equation}

Where, sign(x) is the signum function returning +1 if $x>0$, 0 if $x=0$ and -1 if $x<0$. $\mathbf{1}(.)$ is the indicator function, equal to 1 if its argument is true, and 0 otherwise.
\newline \par

\noindent{\textbf{MAE:}} It calculates the average absolute difference between predicted values ($\hat{y}$) and actual values (${y}$). Mathematically, 

\begin{equation} \label{eqn:mae}
    MAE = \frac{\sum_{i=1}^{n}|y_i-\hat{y_i}|}{n}
\end{equation}

Where $n$ refers to the sample size. MAE is the loss function that need to be minimized in the process of training a machine learning model. Lower MAE values suggest better model performance.

\section{Implementation Details}

We use simulated data to pre-train the model and fine-tune and test in on the collected real-world dara. We employ T4 GPUs with a batch size of 32. The models are trained using the Adam Optimizer with a learning rate of $1\times10^{-3}$ over 200 epochs. To obtain final results, we average the scores of all the clients. Table \ref{table:hyper-param} contains the values of all other hyperparameter used in the study.

\begin{table}[h!]
    \centering
    \caption{List of all the hyperparameter used in the study. The list encompasses values used in MLP model, federated learning and mobile application development (in order).}
    \label{table:hyper-param}
    \begin{tabular}{p{11cm}c}
        \toprule
        \textbf{Hyperparameter} & \textbf{Value}\\
        \midrule
       Batch Size & 32 \\
       Learning Rate & $1\times10^{-3}$ \\
       Pre-training Epochs & 200 \\
       Number of Units in Hidden Layer & [64, 32, 16, 8, 4] \\
       Ideal Sleeping Range & [7 hours, 9 hours] \\
       Ideal Distance Traveled & [5 km, 8 km] \\
       $w_{sleep}$ & 1 \\
       $w_{distance}$ & 0.8 \\
       $w_{bmi}$ & 1 \\
       $w_{meal}$ & 1 \\
       $R_{high}$ & 3 \\
       $\theta$ & 0.5 \\
       Number of Recommendations ($top\_n$) & 2 \\
       \midrule
       Fine-tuning Epochs & 10 \\
       Number of Clients & 10 \\
       \midrule
       $\tau$ & 1.8 \\
       $\delta$ & 0.5m \\
       Sleep Window & [10 PM, 10 AM] \\
        \bottomrule
    \end{tabular}
\end{table}

\let\cleardoublepage\clearpage

\chapter{Results and Discussion}

The results for FedAvg fine-tuning approach and FedPer approach are presented by Table \ref{table:fedavg} and Table \ref{table:fedper}, respectively. It is evident that FedAvg outperforms the FedPer approach. FedAvg achieves an accuracy of 60.71\% and MAE of 0.91 ($\downarrow 0.28$ compared to FedPer). It achieves a 14.37\% more accurate predictions compared to FedPer algorithm.

\begin{table}[h!]
    \centering
    \caption{Client‐wise accuracy (\%) and mean absolute error for the FedAvg approach.}
    \label{table:fedavg}
    \begin{tabular}{p{4cm}cc}
        \toprule
        \textbf{Client ID} & \textbf{Accuracy (\%)} & \textbf{Mean Absolute Error}  \\
        \midrule
       Client 0 & 81.42 & 0.82 \\
        Client 1 & 67.14 & 0.79 \\
        Client 2 & 34.63 & 0.81 \\
        Client 3 & 65.67 & 0.62 \\
        Client 4 & 25.18 & 0.90 \\
        Client 5 & 74.28 & 0.87 \\
        Client 6 & 74.28 & 0.41 \\
        Client 7 & 38.57 & 1.91 \\
        Client 8 & 69.28 & 1.32 \\
        Client 9 & 76.67 & 0.69 \\
        \midrule
        \textbf{Average} & \textbf{60.71} & \textbf{0.91} \\
        \bottomrule
    \end{tabular}
\end{table}

\begin{table}[h!]
    \centering
    \caption{Client‐wise accuracy (\%) and mean absolute error for the FedPer approach.}
    \label{table:fedper}
    \begin{tabular}{p{4cm}cc}
        \toprule
        \textbf{Client ID} & \textbf{Accuracy (\%)} & \textbf{Mean Absolute Error}  \\
        \midrule
       Client 0 & 63.57 & 1.26 \\
        Client 1 & 47.32 & 2.48 \\
        Client 2 & 31.33 & 0.12 \\
        Client 3 & 60.72 & 1.52 \\
        Client 4 & 25.71 & 0.29 \\
        Client 5 & 25.91 & 1.46 \\
        Client 6 & 60.83 & 0.97 \\
        Client 7 & 46.67 & 1.14 \\
        Client 8 & 47.14 & 1.24 \\
        Client 9 & 54.16 & 1.47 \\
        \midrule
        \textbf{Average} & \textbf{46.34} & \textbf{1.19} \\
        \bottomrule
    \end{tabular}
\end{table}

FedAvg exhibits lower accuracy on Clients 2, 4, and 7. These clients’ data are dominated by ideal case (most of their true deficits are zero) - so the model, when averaged across all participants, struggles to reproduce this all‑zero pattern. In effect, the global update dilutes each client’s local bias toward the ideal range, and the resulting model under‑predicts the absence of a deficit. Conversely, FedAvg is better at identifying non‑ideal behavior: for example, Client 0—whose data include a larger proportion of true deficits—achieves the highest accuracy (81.42\%) under the same fine‑tuning regime. This discrepancy highlights two key issues:

\begin{enumerate}
    \item When a client’s labels are majorly zero, the global model—trained on a more heterogeneous mix of deficits fails to specialize on the no deficit case.

    \item FedAvg merges all client updates indiscriminately, which benefits common patterns (e.g. predicting deficits) but washes out client‑specific biases toward ideal behavior.
\end{enumerate}

Together, these factors suggest that purely averaging model weights may not adequately capture per‑client idiosyncrasies, especially for users whose lifestyle metrics remain within recommended ranges.
\newline \par

FedPer yields a slight improvement in accuracy for clients whose data are dominated by ideal zero‐deficit observations, but it conversely underperforms on predicting non‐ideal behaviors. By isolating the last two layers as client‐specific heads, FedPer allows each participant to specialize on their all‐zero patterns—hence the slight gain for ideal cases—yet this same decoupling limits the transfer of deficit‐prediction expertise for rarer non‐ideal events.
\newline \par

This behavior can be explained by:

\begin{enumerate}
  \item Personalized heads faithfully reproduce the zero‐deficit bias present in many clients, driving up ideal‐case accuracy.
  
  \item Since deficit‐related weights in the personal head are never shared, clients with mixed or skewed distributions fail to benefit from peers’ learning on non‑ideal patterns, resulting in lower accuracy on true deficits.
\end{enumerate}

These findings underline the trade‐off inherent in personalization: preserving local idiosyncrasies can boost performance on dominant patterns while potentially impairing the modeling of less frequent but clinically important non‐ideal behaviors.

\let\cleardoublepage\clearpage

\chapter{Conclusion}

In this work, we presented RiM (Record, Improve, and Maintain), a privacy‐preserving mobile application for promoting student physical well–being by combining federated learning with a lightweight MLP recommendation engine. We first pre‐trained our MLP on a large, simulated dataset to learn general patterns of sleep and walking‐distance deficits, then fine‐tuned it on real‐world data collected from IISER Bhopal students using both FedAvg and FedPer strategies. Our experimental results demonstrated that FedAvg achieves higher accuracy on non‐ideal behaviors (average accuracy 60.7\% and MAE 0.91), whereas FedPer better captures client‐specific ideal zero‐deficit patterns (average accuracy 46.3\%, MAE 1.19). To deliver actionable guidance, we devised a hybrid recommendation layer that combines these learned deficits with rule‐based severity and interaction checks, computing composite risk scores to prioritize the top two most urgent, personalized recommendations. 
\newline \par

This approach ensures differential privacy by never transmitting raw user data; model personalization via per‐client heads (FedPer) or global aggregation (FedAvg) as appropriate; and actionable insights through a priority‐based, rule‐enhanced recommender that surfaces only the most critical lifestyle adjustments. However, one \textbf{limitation} of the current RiM implementation is that the Android application supports only Android version 13 or lower, which restricts compatibility with newer devices.
\newline \par

\noindent \textbf{Future Work:} To broaden usability and enhance on‐device intelligence, we will first update the mobile application to support all modern Android API levels. Next, we plan to integrate the pretrained MLP model directly within the app—enabling real‐time, offline inference without server communication. Finally, we will investigate advanced bi‐level optimization algorithms such as Ditto for federated learning, which balance global aggregation with per‐client personalization more effectively, further improving both fairness and predictive performance.

\let\cleardoublepage\clearpage

\small{
\bibliographystyle{unsrt}
\bibliography{references.bib}
\addcontentsline{toc}{chapter}{Bibliography}
}





\end{document}